\documentclass[11pt,a4paper]{article}
\pdfoutput=1
\usepackage[hyperref]{acl2019}
\usepackage{times}
\usepackage{latexsym}
\usepackage{amssymb}
\usepackage{amsmath}

\usepackage{url}

\aclfinalcopy 
\def\aclpaperid{2576} 
\title{A Surprisingly Robust Trick for the Winograd Schema Challenge}

\author{Vid Kocijan\textsuperscript{1}, Ana-Maria Cre\c{t}u\textsuperscript{2}, Oana-Maria Camburu\textsuperscript{1,3}, Yordan Yordanov\textsuperscript{1}, Thomas Lukasiewicz\textsuperscript{1,3} \\
  \textsuperscript{1}University of Oxford \\
  \textsuperscript{2}Imperial College London \\
  \textsuperscript{3}Alan Turing Institute, London \\
  \texttt{firstname.lastname@cs.ox.ac.uk, a.cretu@imperial.ac.uk}}

\date{}

\begin{document}

\maketitle

\begin{abstract}
The Winograd Schema Challenge (\textsc{Wsc}) da\-ta\-set \textsc{Wsc273} and its inference counterpart \textsc{Wnli} are popular benchmarks for natural language understanding and commonsense reasoning.
In this paper, we show that the performance of three language models on \textsc{Wsc273} consistently and robustly improves when fine-tuned on a similar pronoun disambiguation problem dataset (denoted \textsc{WscR}).
We additionally generate a large unsupervised \textsc{Wsc}-like dataset.
By fine-tuning the \textsc{Bert} language model both on the introduced and on the \textsc{WscR} dataset, we achieve overall accuracies of $72.5\%$ and $74.7\%$ on \textsc{Wsc273} and \textsc{Wnli}, improving the previous state-of-the-art solutions by $8.8\%$ and $9.6\%$, respectively. Furthermore, our fine-tu\-ned models are also consistently more accurate on the ``complex'' subsets of \textsc{Wsc273}, introduced by \citeauthor{WSCAnalysis}~(\citeyear{WSCAnalysis}). 
\end{abstract}

\section{Introduction}
\label{Section-introduction}
The Winograd Schema Challenge (\textsc{Wsc}) \cite{levesque2012winograd, WinogradSchema}  was introduced for testing AI agents for commonsense knowledge.
Here, we refer to the most popular collection of such sentences as \textsc{Wsc273}, to avoid confusion with other slightly modified datasets, such as \textsc{Pdp60}, \cite{WSC2016} and the Definite Pronoun Resolution dataset \cite{DPR}, 
denoted \textsc{WscR} in the sequel.
\textsc{Wsc273} consists of 273 instan\-ces of the pronoun disambiguation problem (\textsc{Pdp}) \cite{PlanningWSC}. Each is a sentence (or two) with a pronoun referring to one of the two or more nouns; the goal is to predict the correct one. The task is challenging, since \textsc{Wsc} examples are constructed to require human-like commonsense knowledge and reasoning. 
The best known solutions use deep learning with an accuracy of $63.7\%$ \cite{WSCRanking, WinogradGoogle}. 
The problem is difficult to solve not only because of the commonsense reasoning challenge, but also due to the small existing datasets making it difficult to train neural networks directly on the task.

Neural networks have proven highly effective in natural language processing (NLP) tasks, outperforming other machine learning methods and even matching human performance \cite{human_parity_NMT,HumanGlue}. However, supervised models require many per-task annotated training examples for a good performance. For tasks with scarce data, transfer learning is often applied \cite{Fine-tunedLMs,CNNforText}, i.e., a model that is already trained on one NLP task is used as a starting point for other NLP tasks.

A common approach to transfer learning in NLP is to train a language model (LM) on large amounts of unsupervised text \cite{Fine-tunedLMs} and use it, with or without further fine-tu\-ning, to solve other downstream tasks. Building on top of a LM has proven to be very successful, producing state-of-the-art (SOTA) results \cite{BigBird,WinogradGoogle} on benchmark datasets like \textsc{Glue} \cite{Glue} or \textsc{Wsc273} \cite{WinogradSchema}. 

In this work, we first show that fine-tuning existing LMs on \textsc{WscR}  is a robust method of improving the capabilities of the LM to tackle \textsc{Wsc273} and \textsc{Wnli}.
This is surprising, because previous attempts to generalize from the \textsc{WscR} dataset to \textsc{Wsc273} did not achieve a major improvement \cite{WSCRanking}.
Secondly, we introduce a method for generating large-scale \textsc{Wsc}-like examples.
We use this method to create a $2.4$M dataset 
from  English Wikipedia\footnote{\url{https://dumps.wikimedia.org/enwiki/} dump id: enwiki-20181201}, which we further use together with  \textsc{WscR}  for fine-tuning the pre-trained \textsc{Bert} LM \cite{Bert}.
The dataset and the code have been made publicly available\footnote{The code can be found at \url{https://github.com/vid-koci/bert-commonsense}.\\ The dataset and the models can be obtained from \url{https://ora.ox.ac.uk/objects/uuid:9b34602b-c982-4b49-b4f4-6555b5a82c3d}}.
We achieve accuracies of $72.5\%$ and $74.7\%$ on \textsc{Wsc273}  and \textsc{Wnli}, improving the previous best solutions by $8.8\%$ and $9.6\%$, respectively.

\section{Background}\label{Section:4}
This section introduces the main LM used in our work, \textsc{Bert} \cite{Bert}, followed by a detailed description of \textsc{Wsc} and its relaxed form, the Definite Pronoun Resolution problem.

\paragraph*{\textsc{Bert}.}
Our work uses the pre-trained Bidirectional Encoder Representations from Transformers (\textsc{Bert}) LM \cite{Bert} based on the transformer architecture \cite{AttentionAllYouNeed}.
Due to its high performance on natural language understanding (NLU) benchmarks and the simplicity to adapt its objective function to our fine-tuning needs, we use \textsc{Bert} throughout this work.

\textsc{Bert} is originally trained on two tasks: masked token prediction, where the goal is to predict the missing tokens from the input sequence, and next sentence prediction, where the model is given two sequences and asked to predict whether the second sequence follows after the first one.

We focus on the first task to fine-tune \textsc{Bert} using \textsc{Wsc}-like examples.
We use masked token prediction on a set of sentences that follow the \textsc{Wsc} structure, where we aim to determine which of the candidates is the correct replacement for the masked pronoun.

\paragraph*{Winograd Schema Challenge.} \label{WSCsection}
Having introdu\-ced the goal of the Winograd Schema Challenge in Section~\ref{Section-introduction}, we illustrate it with the following example:

\smallskip 
\textit{The trophy didn't fit into the suitcase because \textbf{it} was too [large/small].}

\textbf{Question: } What was too [large/small]?

\textbf{Answer: } the trophy / the suitcase

\smallskip 
The pronoun ``\textbf{it}'' refers to a different noun, based on the word in the brackets.
To correct\-ly answer both versions, one must understand the meaning of the sentence and its relation to the changed word.
More specifically, a text must meet the following criteria to be considered for a Winograd Schema \cite{WinogradSchema}:
\begin{enumerate}
 \vspace*{-1.5ex}   \item Two parties must appear in the text.
 \vspace*{-1.5ex}   \item A pronoun or a possessive adjective appears in the sentence and refers to one party. It would be grammatically correct if it referred to the other.
 \vspace*{-1.5ex}   \item The question asks to determine what party the pronoun or the possessive adjective refers to.
 \vspace*{-1.5ex}   \item A ``special word'' appears in the sentence. When switched to an ``alternative word'', the sentence remains grammatically correct, but the referent of the pronoun changes.
\end{enumerate}

Additionally, commonsense reasoning must be required to answer the question.



A detailed analysis by \citeauthor{WSCAnalysis}~(\citeyear{WSCAnalysis}) shows that not all \textsc{Wsc273} examples are equally difficult.
They introduce two complexity measures (associativity and switchability) and, based on them, refine evaluation metrics for  \textsc{Wsc273}.

In \textit{associative} examples, one of the parties is more commonly associated with the rest of the question than the other one.
Such examples are seen as ``easier'' than the rest and represent $13.5\%$ of \textsc{Wsc273}.
The remaining $86.5\%$ of \textsc{Wsc273} is called \textit{non-associative}.

$47\%$ of the examples are ``switchable'', because the roles of the parties can be changed, and examples still make sense.
A model is tested on the original, ``unswitched'' switchable subset and on the same subset with switched parties.
The consistency between the two results is computed by comparing how often the model correctly changes the answer when the parties are switched.

\paragraph*{Definite Pronoun Resolution.}

Since collecting examples that meet the criteria for \textsc{Wsc} is hard, \citeauthor{DPR}~(\citeyear{DPR}) relax the criteria and construct the Definite Pronoun Resolution (\textsc{Dpr}) dataset, following the structure of  \textsc{Wsc}, but also accepting easier examples. 
The dataset, referred throughout the paper as \textsc{WscR}, is split into a training set with $1322$ examples and test set with $564$ examples. 
Six examples in the \textsc{WscR} training set reappear in \textsc{Wsc273}.
We remove these examples from \textsc{WscR}.
We use the \textsc{WscR} training and test sets for fine-tuning the LMs and for validation, respectively.

\paragraph{\textsc{Wnli}.}
One of the 9 \textsc{Glue} benchmark tasks \cite{Glue}, \textsc{Wnli} is very similar to the  \textsc{Wsc273} dataset, but is phrased as an entailment problem instead.
A \textsc{Wsc} schema is given as a premise. The hypothesis is constructed by extracting the sentence part where the pronoun is, and replacing the pronoun with one candidate. The label is 1, if the candidate is the correct replacement, and 0, otherwise.

\section{Related Work}

There have been several attempts at solving  \textsc{Wsc273}.  
Previous work is based on Google queries for knowledge \cite{KnowledgeHunter} ($58\%$), sequence ranking \cite{WSCRanking} ($63\%$), and using an ensemble of LMs \cite{WinogradGoogle} ($63\%$).


A critical analysis \cite{WSCAnalysis} showed that the main reason for success when using an ensemble of LMs \cite{WinogradGoogle} was largely due to imperfections in \textsc{Wsc273}, as discussed in Section \ref{Section:4}.

The only dataset similar to \textsc{Wsc273} is an easier but larger ($1886$ examples) variation published by \citeauthor{DPR}~(\citeyear{DPR}) and earlier introduced as \textsc{WscR}.
The sequence ranking approach uses \textsc{WscR} for training and attempts to generalize to \textsc{Wsc273}.
The gap in performance scores between \textsc{WscR} and \textsc{Wsc273} ($76\%$ vs.~$63\%$) implies that examples in \textsc{Wsc273} are much harder.
We note that \citeauthor{WSCRanking}~(\citeyear{WSCRanking}) do not report removing the overlapping examples between \textsc{WscR} and \textsc{Wsc273}.

Another important NLU benchmark is \textsc{Glue} \cite{Glue}, which gathers $9$ tasks and is commonly used to evaluate LMs.
The best score has seen a huge jump from $0.69$ to over $0.82$ in a single year.
However, \textsc{Wnli} is a notoriously difficult task in \textsc{Glue} and remains unsolved by the existing approaches.
None of the models have beaten the majority baseline at $65.1$, while human performance lies at $95.9$ \cite{HumanGlue}.
\section{Our Approach}\label{Section:5}

\paragraph{\textsc{Wsc} Approach.}
We approach \textsc{Wsc} by fine-tuning the pre-trained \textsc{Bert} LM \cite{Bert} on the \textsc{WscR} training set and further on a very large Winograd-like dataset that we introduce. 
Below, we present our fine-tuning objective function and the introduced dataset.

Given a training sentence $\mathbf{s}$, the pronoun to be resolved is masked out from the sentence, and the LM is used to predict the correct candidate in the place of the masked pronoun.
Let $c_1$ and $c_2$ be the two candidates.
\textsc{Bert} for Masked Token Prediction is used to find $\mathbb{P}(c_1|\mathbf{s})$ and $\mathbb{P}(c_2|\mathbf{s})$.
If a candidate consists of several tokens, the corresponding number of \texttt{[MASK]} tokens is used in the masked sentence.
Then, $\log\mathbb{P}(c|\mathbf{s})$ is computed as the average of log-probabilities of each composing token. 
If $c_1$ is correct, and $c_2$ is not, the loss is:
\begin{align}\label{loss}
L=-&\log\mathbb{P}(c_1|\mathbf{s})\ +\\
+\ &\alpha\cdot\text{max}(0,\log\mathbb{P}(c_2|\mathbf{s})-\log\mathbb{P}(c_1|\mathbf{s})+\beta), \nonumber
\end{align}%
where $\alpha$ and $\beta$ are hyperparameters.

\paragraph*{MaskedWiki Dataset.} %
To get more data for fine-tuning, we automatically generate a large-scale collection of sentences similar to \textsc{Wsc}.
More specifically, our procedure searches a large text corpus for sentences that contain (at least) two occurrences of the same noun.
We mask the second occurrence of this noun with the \texttt{[MASK]} token. 
Several possible replacements for the masked token are given, for each noun in the sentence different from the replaced noun.
We thus obtain examples that are structurally similar to those in \textsc{Wsc}, although we cannot ensure that they fulfill all the requirements (see Section \ref{Section:4}).

To generate such sentences, we choose the English Wikipedia as source text corpus, as it is a large-scale and grammatically correct collection of text with diverse information. 
We use the Stanford POS tagger \cite{StanfordPOS} for finding nouns.
We obtain a dataset with approximately $130$M examples.
We downsample the dataset uniformly at random to obtain a dataset of manageable size.
After downsampling, the dataset consists of $2.4$M examples.
All experiments are conducted with this downsampled dataset only.

To determine the quality of the dataset, $200$ random examples are manually categorized into $4$ categories:
\begin{itemize}
  \vspace*{-1ex}  \item Unsolvable: the masked word cannot be unambiguously selected with the given context. Example:
    \textit{Palmer and Crenshaw both used Wilson 8802 putters , with [MASK] 's receiving the moniker  `` Little Ben '' due to his proficiency with it . [Palmer/Crenshaw]}
  \vspace*{-1ex}  \item Hard: the answer is not trivial to figure out. Example: \textit{At the time of Plath 's suicide , Assia was pregnant with Hughes 's child , but she had an abortion soon after [MASK] 's death . [Plath/Assia]}
  \vspace*{-1ex}  \item Easy: The alternative sentence is grammatically incorrect or is very visibly an inferior choice. Example: \textit{The syllables are pronounced strongly by Gaga in syncopation while her vibrato complemented Bennett's characteristic jazz vocals and swing . Olivier added , `` [MASK] 's voice , when stripped of its bells and whistles, showcases a timelessness that lends itself well to the genre~.~'' [Gaga/syncopation]}
  \vspace*{-1ex}  \item Noise: The example is a result of a parsing error.
\end{itemize}
In the analyzed subset, $8.5\%$ of examples were unsolvable, $45\%$ were hard, $45.5\%$ were easy, and $1\%$ fell into the noise category.

\paragraph*{\textsc{Wnli} Approach.}
Models are additionally tested on the test set of the \textsc{Wnli} dataset.
To use the same evaluation approach as for the \textsc{Wsc273} dataset, we transform the examples in \textsc{Wnli} from the premise--hypothesis format into the masked words format.
Since each hypothesis is just a substring of the premise with the pronoun replaced for the candidate, finding the replaced pronoun and one candidate can be done by finding the hypothesis as a substring of the premise.
All other nouns in the sentence are treated as alternative candidates.
The Stanford POS-tagger \cite{StanfordPOS} is used to find the nouns in the sentence.
The probability for each candidate is computed to determine whether the candidate in the hypothesis is the best match.
Only the test set of the \textsc{Wnli} dataset is used, because it does not overlap with \textsc{Wsc273}.
We do not train or validate on the \textsc{Wnli} training and validation sets, because some of the examples share the premise. Indeed, when upper rephrasing of the examples is used, the training, validation, and test sets start to overlap.

\section{Evaluation}\label{Evaluation}

\begin{table*}[ht!]
\begin{tabular}{@{\,}l@{\ \ \ }c@{\ \ \ }c@{\ \ \ }c@{\ \ \ }c@{\ \ \ }c@{\ \ \ }c@{\ \ \ }c@{\,}}
 & \textsc{Wsc273} & non-assoc. & assoc. & unswitched & switched & consist. & WNLI  \\\hline
\textsc{Bert\_Wiki} & $0.619$ & $0.597$ & $0.757$ & $0.573$ & $0.603$ & $0.389$ & $0.712$  \\
\textsc{Bert\_Wiki\_WscR} & $\mathbf{\underline{0.725}}$ & $\mathbf{\underline{0.720}}$ & $\mathbf{0.757}$ & $\mathbf{\underline{0.732}}$ & $\mathbf{\underline{0.710}}$ & $\mathbf{0.550}$ & $\underline{\mathbf{0.747}}$ \\ \hline
\textsc{Bert} & $0.619$ & $0.602$ & $0.730$ & $0.595$ & $0.573$ & $0.458$ & $0.658$ \\
\textsc{Bert\_WscR} & $\mathbf{0.714}$ & $\mathbf{0.699}$ & $\mathbf{0.811}$ & $\mathbf{0.695}$ & $\mathbf{0.702}$ & $\mathbf{0.550}$ & $\mathbf{0.719}$  \\\hline
\textsc{Bert}-base & $0.564$ & $0.551$ & $0.649$ & $0.527$ & $0.565$ & $0.443$ &  $0.630$\\
\textsc{Bert}-base\_\textsc{WscR} & $\mathbf{0.623}$ & $\mathbf{0.606}$ & $\mathbf{0.730}$ & $\mathbf{0.611}$ & $\mathbf{0.634}$ & $\mathbf{0.443}$ & $\mathbf{0.705}$  \\\hline
\textsc{Gpt} & $0.553$ & $0.525$ & $0.730$ & $0.595$ & $0.519$ & $0.466$ & -- \\
\textsc{Gpt\_WscR} & $\mathbf{0.674}$ & $\mathbf{0.653}$ & $\mathbf{0.811}$ & $\mathbf{0.664}$ & $\mathbf{0.580}$ & $\mathbf{0.641}$ & --  \\\hline
\textsc{Bert\_Wiki\_WscR}\_no\_pairs & $0.663$ & $0.669$ & $0.622$ & $0.672$ & $0.641$ & $0.511$ & -- \\
 \textsc{Bert\_Wiki\_WscR}\_pairs& $\mathbf{0.703}$ & $\mathbf{0.695}$ & $\mathbf{0.757}$ & $\mathbf{0.718}$ & $\mathbf{0.710}$ & $\mathbf{0.565}$ & -- \\\hline
LM ensemble & $0.637$ & $0.606$ & $\underline{\mathbf{0.838}}$ & $0.634$ & $0.534$ & $0.443$ & -- \\
Knowledge Hunter & $0.571$ & $0.583$ & $0.5$ & $0.588$ & $0.588$ & $\underline{\mathbf{0.901}}$ & --
\end{tabular}
\caption{Results on \textsc{Wsc273} and its subsets. The comparison between each language model and its \textsc{WscR}-tuned model is given. For each column, the better result of the two is in bold. The best result in the column overall is underlined. Results for the LM ensemble and Knowledge Hunter are taken from \citeauthor{WSCAnalysis}~(\citeyear{WSCAnalysis}). All models consistently improve their accuracy when fine-tuned on the \textsc{WscR} dataset.\vspace*{-3ex}}
\label{results}
\end{table*}

In this work, we use the PyTorch implementation\footnote{\url{https://github.com/huggingface/pytorch-pretrained-BERT}} of \citeauthor{Bert}'s (\citeyear{Bert}) pre-trained model, \textsc{Bert}-large. 
To obtain \textsc{Bert\_Wiki}, we train on MaskedWiki starting from the pre-trained \textsc{Bert}.
The training procedure differs from the training of \textsc{Bert} \cite{Bert} in a few points.
The model is trained with a single epoch of the MaskedWiki dataset, using batches of size $64$  (distributed on $8$ GPUs), Adam optimizer, a learning rate of $5.0\cdot 10^{-6}$, and hyperparameter values $\alpha=20$ and $\beta=0.2$ in the loss function (Eq.~(\ref{loss})). The values were selected from $\alpha\in\{5,10,20\}$ and $\beta\in\{0.1,0.2,0.4\}$ and learning rate from $\{3\cdot 10^{-5},1\cdot 10^{-5},5\cdot 10^{-6},3\cdot 10^{-6}\}$ using grid search.
To speed up the hyperparameter search, the training (for hyperparameter search only) is done on a randomly selected subset of size $100,000$. 
The performance is then compared on the \textsc{WscR} test set.


Both \textsc{Bert} and \textsc{Bert\_Wiki} are fine-tuned on the \textsc{WscR} training dataset to create \textsc{Bert\_WscR} and \textsc{Bert\_Wiki\_WscR}.

The \textsc{WscR} test set was used as the validation set.
The fine-tuning procedure was the same as the training procedure on MaskedWiki, except that $30$ epochs were used.
The model was validated after every epoch, and the model with highest performance on the validation set was retained.
The hyperparameters $\alpha$ and $\beta$ and learning rate were selected with grid search from the same sets as for MaskedWiki training.

For comparison, experiments are also conducted on two other LMs, \textsc{Bert}-base (\textsc{Bert} with less parameters) and General Pre-trained Transformer (\textsc{Gpt}) by \citeauthor{GPT}~(\citeyear{GPT}).
The training on \textsc{Bert}-base was conducted in the same way as for the other models.
When using \textsc{Gpt}, the probability of a word belonging to the sentence $\mathbb{P}(c|\mathbf{s})$ is computed as partial loss in the same way as by \citeauthor{WinogradGoogle}~(\citeyear{WinogradGoogle}).

Due to \textsc{Wsc}'s ``special word'' property, examples come in pairs.
A pair of examples only differs in a single word (but the correct answers are different).
The model \textsc{Bert\_Wiki\_WscR}\_no\_pairs is the \textsc{Bert\_Wiki} model, fine-tuned on \textsc{WscR}, where only a single example from each pair is retained.
The size of \textsc{WscR} is thus halved.
The model \textsc{Bert\_Wiki\_WscR}\_pairs is obtained by fine-tuning \textsc{Bert\_Wiki}  on half of the \textsc{WscR} dataset.
This time, all examples in the subset come in pairs, just like in the unreduced \textsc{WscR} dataset.

We evaluate all models on \textsc{Wsc273} and the \textsc{Wnli} test dataset, as well as the various subsets of \textsc{Wsc273}, as described in Section \ref{Section:4}. The results are reported in Table \ref{results} and will be discussed next.
\paragraph*{Discussion.}
Firstly, we note that models that are fine-tuned on the \textsc{WscR} dataset consistently outperform their non-fine-tuned counterparts.
The \textsc{Bert\_Wiki\_WscR} model outperforms other language models on $5$ out of $6$ sets that they are compared on.
In comparison to the LM ensemble by \citeauthor{WinogradGoogle}~(\citeyear{WinogradGoogle}), the accuracy is more consistent between associative and non-associative subsets and less affected by the switched parties.
However, it remains fairly inconsistent, which is a general property of LMs.

Secondly, the results of \textsc{Bert\_Wiki} seem to indicate that this dataset alone does not help \textsc{Bert}. However, when additionally fine-tuned to \textsc{WscR}, the accuracy consistently improves.

Finally, the results of \textsc{Bert\_Wiki}\_no\_pairs and \textsc{Bert\_Wiki}\_pairs show that the existence of \textsc{Wsc}-like pairs in the training data affects the performance of the trained model.
MaskedWiki does not contain such pairs.

\section{Summary and Outlook}
This work achieves new SOTA results on the \textsc{Wsc273} and \textsc{Wnli} datasets by fine-tuning the \textsc{Bert} language model on the \textsc{Wscr} dataset and a newly introduced MaskedWiki dataset.
The previous SOTA results on \textsc{Wsc273} and \textsc{Wnli} are improved by $8.8\%$ and $9.6\%$, respectively.
To our knowledge, this is the first model that beats the majority baseline on \textsc{Wnli}.

We show that by fine-tuning on \textsc{Wsc}-like data, the language model's performance on \textsc{Wsc} consistently improves.
The consistent improvement of several language models indicates the robustness of this method.
This is particularly surprising, because previous work \cite{WSCRanking} implies that generalizing to \textsc{Wsc273} is hard.

In future work, other uses and the statistical significance of MaskedWiki's impact and its applications to different tasks will be investigated.
Furthermore, to further improve the results on \textsc{Wsc273}, data-filtering procedures may be introduced to find harder \textsc{Wsc}-like examples.

\section*{Acknowledgments}
This work was supported  by the Alan Turing Institute under the UK EPSRC grant EP/N510129/1, by the EPSRC grant EP/R013667/1, by the EPSRC studentship OUCS/EPSRC-NPIF/VK/ 1123106, and 
by an EPSRC Vacation Bursary. 
We also acknowledge the use of the EPSRC-funded Tier 2 facility JADE (EP/P020275/1).

\bibliographystyle{acl_natbib}
\bibliography{refs}

\end{document}